\pdfoutput=1
\documentclass[11pt]{article}
\usepackage{coling2020}
\usepackage{times}
\usepackage{url}
\usepackage{latexsym}
\usepackage{graphicx}
\usepackage{lineno}
\usepackage{amssymb}
\usepackage{amsmath}
\usepackage{booktabs}
\usepackage{bm}
\usepackage{color}
\usepackage{enumitem}
\usepackage{appendix}
\setlist{nosep}
\newcommand{\tabincell}[2]{\begin{tabular}{@{}#1@{}}#2\end{tabular}}

\colingfinalcopy 

\title{\textsc{Chime}: Cross-passage Hierarchical Memory Network for Generative Review Question Answering}

\author{Junru Lu$^1$, Gabriele Pergola$^1$, Lin Gui$^1$, Binyang Li$^2$ and Yulan He$^1$ \\
  $^1$Department of Computer Science, University of Warwick, UK \\
  $^2$School of Information Science and Technology, \\University of International Relations, Beijing, China\\
    {\tt \{Junru.Lu, Gabriele.Pergola, Lin.Gui, Yulan.He\}@warwick.ac.uk} \\
  \tt {byli@uir.edu.cn}}

\date{}

\begin{document}
\maketitle
\begin{abstract}
  We introduce \textsc{Chime}, a cross-passage hierarchical memory network for question answering (QA) via text generation. It extends XLNet \cite{yang2019xlnet} introducing an auxiliary memory module consisting of two components: the \textit{context memory} collecting cross-passage evidence, and the \textit{answer memory} working as a buffer continually refining the generated answers. Empirically, we show the efficacy of the proposed architecture in the multi-passage generative QA, outperforming the state-of-the-art baselines with better syntactically well-formed answers and increased precision in addressing the questions of the AmazonQA review dataset. An additional qualitative analysis revealed the interpretability introduced by the memory module\blfootnote{This work is licensed under a Creative Commons Attribution 4.0 International Licence. Licence details: \url{http://creativecommons.org/licenses/by/4.0/}.}.
\end{abstract}

\section{Introduction}
With the development of large-scale pre-trained Language Models (LMs) such as BERT \cite{devlin2018bert}, XLNet \cite{yang2019xlnet}, and T5 \cite{raffel2019exploring}, tremendous progress has been made in Question Answering (QA). Fine tuning pre-trained LMs on task-specific data has surpassed human performance on QA datasets such as SQuAD \cite{rajpurkar2016squad} and NewsQA \cite{trischler2016newsqa}. 
Nevertheless, most existing QA systems largely deal with factoid questions and assume a simplified setup such as multiple-choice questions, retrieving spans of text from given documents, and filling in the blanks. However, in many more realistic situations such as online communities, people tend to ask ‘\emph{descriptive}’ questions (e.g., ‘\emph{How to improve the sound quality of echo dot?}’). Answering such questions requires the identification, linking, and integration of relevant information scattered over long-form multiple documents for the generation of free-form answers.

We are particularly interested in developing a QA system for questions from e-shopping communities using customer reviews. Compared to factoid QA systems, building a review QA system faces the following challenges: (1) as opposed to extractive QA where answers can be directly extracted from documents or multiple-choice QA where systems only need to make a selection over a set of pre-defined answers, review QA needs to gather evidence across multiple documents and generate answers in free-form text; (2) while factoid QA mostly centres on `entities' and only needs to deal with limited types of questions, review QA systems are often presented with a wide variety of ‘\emph{descriptive}’ questions; (3) customer reviews may contain contradictory opinions. Review QA systems need to automatically identify the most prominent opinion given a question for answer generation.

\begin{figure}[ht]
\centering
\includegraphics[scale=0.245]{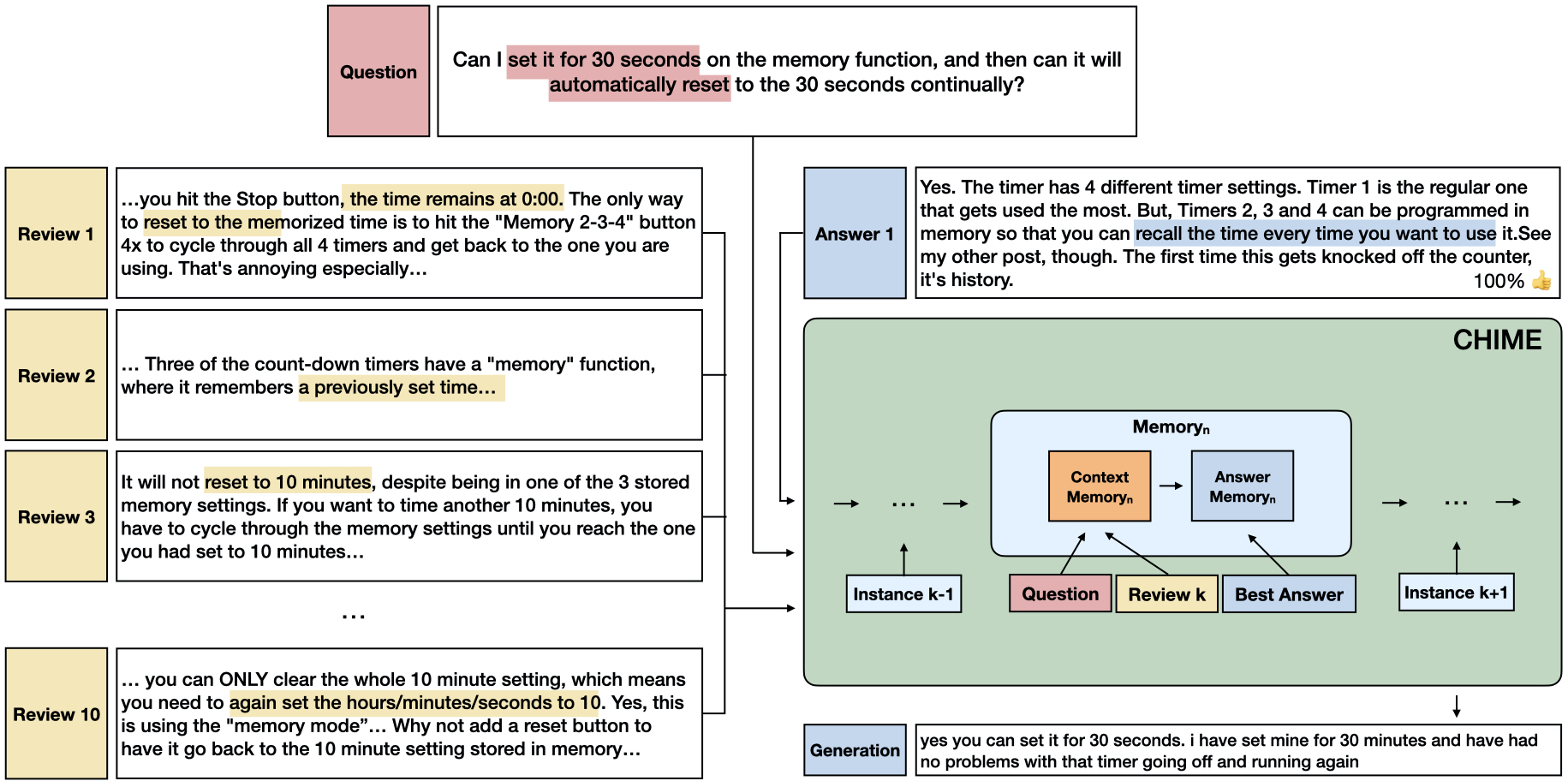}
\caption{Illustration of the review QA task and the general idea of \textsc{Chime}. The example question (the top box) is paired with 10 reviews (left panel) and one or more answers (right upper panel). \textsc{Chime} is trained on the (Question, Review, Answer) triplet. During testing, \textsc{Chime} is presented with a question and 10 related reviews and generates an answer (right bottom box). Both reviews and answers in this example contain contradictory information as highlighted by colors, while the question contains complex sub-questions. \textsc{Chime} is able to identify relevant evidence and generate clear answers.}
\label{fig:illus}
\end{figure}

In our work here, we focus on the AmazonQA dataset \cite{gupta2019amazonqa}, which contains a total of 923k questions and most of the questions are associated with 10 reviews and one or more answers. We propose a novel Cross-passage Hierarchical Memory Network named \textsc{Chime} to address the aforementioned challenges. Regular neural QA models search answers by interactively comparing the question and supporting text, which is in line with human cognition in solving factoid questions \cite{2019Human,1987Literacy}. While for opinion questions, the cognition process is deeper: reading larger scale and more complex texts, building cross-text comprehension, continually refine the opinions, and finally form 
the answers \cite{2019Human}. Therefore, \textsc{Chime} is designed to maintain hierarchical dual memories to closely simulates this cognition process. In this model, a \emph{context memory} dynamically collect cross-passage evidences, an \emph{answer memory} stores and continually refines answers generated as \textsc{Chime} reads supporting text in a sequential manner. Figure \ref{fig:illus} illustrates the setup of our task and an example output generated from \textsc{Chime}. The top box shows a question extracted from our test set while the left panel and the right upper panel show the related 10 reviews and the paired 4 actual answers. We can observe that the question can be decomposed into complex sub-questions and both reviews and answers contain contradictory information. However, \textsc{Chime} can deal with such information effectively and generate appropriate answers as shown in the right-bottom box.

In summary, we have made the following contributions:
\begin{itemize}
\item We propose a novel Cross-passage HIerarchical MEmory Network (\textsc{Chime}) for review QA. Compared with many multi-passage QA models, \textsc{Chime} does not rely on explicit helpful ranking information of supporting reviews, but can capture cross-passage contextual information and effectively identify the most prominent opinion in reviews.
\item \textsc{Chime} reads reviews sequentially, overcoming the input length limitation affecting most of the existing transformer-based systems, and brings some interpretability for these "black box" models.
\item Experimental results on the AmazonQA dataset show that \textsc{Chime} outperforms a number of competitive baselines in terms of the quality of answers generated.
\end{itemize}

\section{Related Work}
Our work is related to the following three lines of research:

\paragraph{Opinion/Review Question-Answering} In Opinion or Review QA, questions may concern about finding subjective personal experiences or opinions of certain products and services. The Amazon QA dataset was first released in \cite{mcauley2016addressing} which contains 1.4 million questions (and answers) and 13 million reviews on 191 thousand products collected from Amazon product pages. They developed a Mixture of Expert (MoE) model which automatically detects whether a review of a product is relevant to a given query. In their subsequent work, Wan and McAuley \shortcite{wan2016modeling} noticed that users tend to ask for \emph{subjective} information and answers might also be highly subjective and possibly contradictory. They, therefore, built a new dataset with 800 thousand questions and over 3 million answers from Amazon, in which each question is paired with multiple answers, and extended their previous MoE model with subjective information such as review rating scores and reviewer's bias incorporated. But they found that subjective information is only effective in predicting `yes/no' answers to binary questions and does not help in distinguishing `true’ answers from alternatives in open-ended '\emph{descriptive}' questions. More recently, Yu and Lam \shortcite{yu2018aware} only focused on the yes/no questions in the Amazon QA dataset \cite{mcauley2016addressing} and trained a binary answer prediction model by leveraging latent aspect-specific representations of both questions and reviews learned by an autoencoder. Gao et al. \shortcite{gao2019product} focused on factual QA in e-commerce and proposed a Product-Aware Answer Generator that combines reviews and product attributes for answer generation, and uses a discriminator to determine whether the generated answer contains question-related facts. Xu et al. \shortcite{xu2019bert} proposed an extractive review-based QA task and manually created just over 2,500 questions and annotated the corresponding answer spans in less than 1,000 reviews relating to laptops and restaurants from the review data of SemEval 2016 Task 5\footnote{\url{http://alt.qcri.org/semeval2016/task5/}}. They first jointly fine-tuned BERT for answer span detection, aspect extraction and aspect sentiment classification on the SemEval 2016 Task 5 data, and then post-trained BERT on over 3 million unlabelled Amazon and Yelp reviews in order to fuse domain knowledge, and also on SQuAD 1.1 \cite{rajpurkar2016squad} in order to gain task-relevant but out-of-domain knowledge. Gupta et al. \shortcite{gupta2019amazonqa} created a subset from the Amazon QA product review dataset \cite{mcauley2016addressing}, consisting of 923k questions with 3.6M answers and 14M reviews on 156k Amazon products. They trained an answerability classifier from 3,297 question-context pairs labeled by Mechanical Turk and used it to classify answerability for the whole dataset. They then converted the dataset into a span-based format by heuristically creating an answer span from reviews that best answers a question based on users' actual answers, and trained R-Net \cite{wang17}, which uses a gated self-attention mechanism and pointer networks, to predict answer boundaries. There are few studies using generative models to deal with opinion/review-based QA.

\paragraph{Multi-passage QA} 
There are mainly two types of methods for multi-passage QA. One is to use retrieval-based methods to first identify text passages that are most likely to contain answer information, and then perform QA on the extracted text passages which are essentially considered as a single passage. The other one is to separately run single-passage QA over each passage, obtaining multiple answer candidates, and then determine the best answer through mutual verification among the answers.

Examples in the first type of methods include S-NET \cite{Tan18}, Multi-passage BERT \cite{wang2019multi}, and Masque \cite{nishida19}. These models require supporting text passages to be explicitly annotated. S-NET \cite{Tan18} follows an extraction-then-synthesis framework. First, relevant passages are extracted from context using a variant of R-NET \cite{wang17}, which learns to rank passages and extract the most possible evidence span from the selective passage; then, the evidence-notated selective passage is used for the GRU decoder synthesizing answers. In Multi-passage BERT \cite{wang2019multi}, two independent BERTs were used to perform multi-passage QA. One BERT takes the question and a text passage as input and then uses the hidden states of the CLS token to train a classifier to determine if the text passage is relevant to the given question. The other BERT is used for extracting candidate answers from relevant text passages. The Masque model \cite{nishida19} is a generative reading comprehension approach based on multi-source abstractive summarization. Masque uses a joint-learning framework, comprising of a question answerability classifier, a passage ranker, and an answer generator. At each step of answer generation, the decoder chooses a word from the mixture of three distributions derived from a vocabulary, from the question and associated multiple passages. 
A representative example of the second type of methods is V-Net \cite{wang2018multi}. The main assumption of V-Net is that correct answers often appear in multiple documents with high frequency and similarity, and wrong answers are usually different from each other. Therefore, V-Net builds a mutual verification mechanism between all answer candidates, which are separately extracted from different passages, to select the best final answer. 

Most existing approaches require explicit annotations of supporting text passages in order to train multi-passage QA models in a supervised way. In our setup here, supporting review passages to a question was unsupervised ranked by BM25, which may introduce noises to QA model training and poses a more significant challenge.

\paragraph{Memory Network} 
Memory network has been first proposed to model the relation between a story and a query for QA systems \cite{weston2014memory,sukhbaatar2015end}. Apart from its application in QA, memory networks have also achieved great successes in other NLP tasks, such as machine translation \cite{maruf2017document}, sentiment analysis \cite{fan2018convolution}, visual question answering \cite{xiong2016dynamic}, social networks \cite{fu2020recurrent}, and summarization \cite{kim2018abstractive}. The main idea of memory networks is to use the attention mechanism to assign different weights to text passages so as to identify the most relevant passages for answer generation \cite{weston2014memory}. Kumar et al.  ~\shortcite{kumar2016ask} proposed a gated memory network to represent facts in different iterations during the learning process to verify the potentially related passages to generate an answer. Gui et al. \shortcite{gui2017question} used a convolutional architecture to capture attention signals in memory networks.  Xu et al.~\shortcite{xu2019enhancing} leveraged the memory network as an information retrieval system to search possible entities in knowledge bases for complex questions. Chen et al.~\shortcite{chen2019bidirectional} used the memory network to verify items in knowledge bases as passages and then generate answers. Generally speaking, existing memory-network-based QA methods mainly focus on using memory networks to weigh and derive representations of question-aware text passages and knowledge entities for answer generation. We instead explore a novel structure of a hierarchical memory network composing of both \emph{context} and \emph{answer} memories for better capturing review context and generating more appropriate answers.

\section{Cross-passage Hierarchical Memory Network (\textsc{Chime})}

In this section, we first define the review QA task and then present our proposed Cross-passage Hierarchical Memory Network (\textsc{Chime}).

\subsection{Task Formulation}
\label{sec:task formulation}

We focus on generative QA with multiple reviews and develop our model based on the AmazonQA dataset \cite{gupta2019amazonqa} in which most of the questions is paired with multiple answers and the top 10 most relevant text snippets as supporting passages extracted from the associated reviews by BM25 \cite{robertson2009probabilistic}. In addition, each question is annotated if it is answerable based on the top 10 review snippets, and each answer is 
accompanied with response votes. The review QA task can be defined as: given an answerable question $\bm{x}^q=\{x_{1}^{q},x_{2}^{q},\cdots,x_{N_q}^{q}\}$, $K$ supporting reviews with $k$-th review represented as $\bm{x}^{r_k}=\{x_{1}^{r_{k}},x_{2}^{r_{k}},\cdots,x_{N_r}^{r_{k}}\}$, a model is asked to generate an answer $\bm{\hat{y}}=\{\hat{y}_{1},
\hat{y}_{2},\cdots,\hat{y}_{N_a}\}$, where $N_q, N_r$ and $N_a$ denote the length of a question, a review and an answer, respectively. In training phase, $L$ answers with $l$-th answer represented as $\bm{y^{a_{l}}}=\{y^{a_{l}}_{1},
y^{a_{l}}_{2},\cdots,y^{a_{l}}_{N_a}\}$ and corresponding response votes $\bm{v^{a_{l}}}=\{v^{a_{l}}_{+},v^{a_{l}}\}$ are provided, where $v^{a_{l}}_{+}$ denotes the number of positive votes and $v^{a_{l}}$ denotes the number of all votes, and $0 \leq v^{a_{l}}_{+} \leq v^{a_{l}}$. 

\subsection{\textsc{Chime}}
\begin{figure}[ht]
\centering
\includegraphics[scale=0.235]{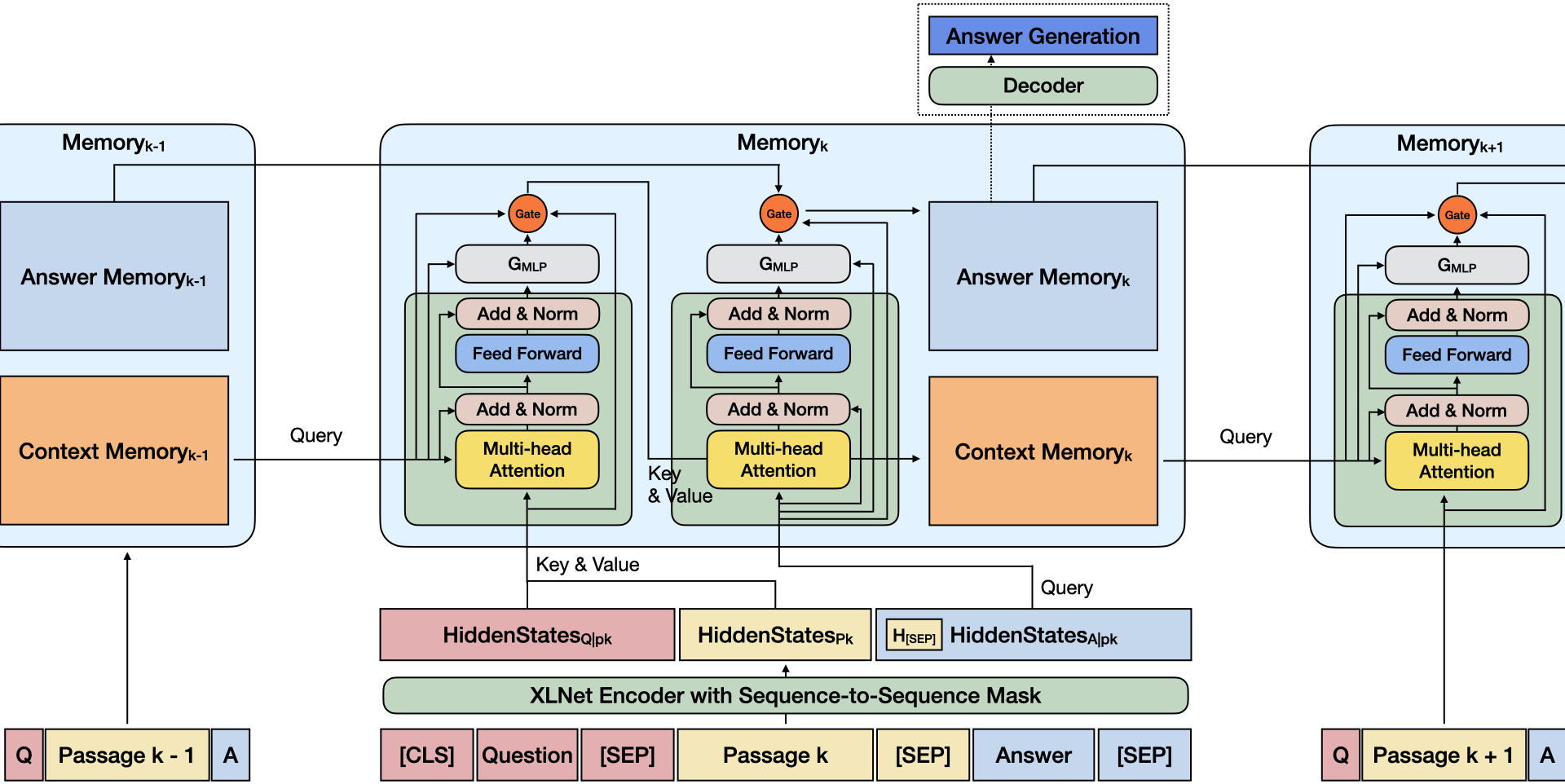}
\caption{The architecture of Cross-passage Hierarchical Memory Network (\textsc{Chime}). The model reads multiple reviews in a sequential manner. When reading the instance $k$ consists of the question, $k$-th review, and the gold answer, the model first derive hidden states of the instance $k$ from the XLNet encoder, then use the hidden states of context part update the \emph{context memory} (the left part of the Memory$_{k}$). With the newly updated context memory, \textsc{Chime} then be able to use the hidden states of the answer part to update the \emph{answer memory} (the middle part of the Memory$_{k}$). After reading the last review, the \emph{answer memory} will be input to the decoder and get a final answer generation (the top dotted frame).}
\label{fig:Arch}
\end{figure}

In this paper, we propose a Cross-passage HIerarchical MEmory Network (\textsc{Chime}) for review question answering. As has been shown in \cite{petroni2019language}, pre-trained LMs can be used as implicit knowledge bases, making them suitable for language generation. Hence, in this paper, we leverage the XLNet \cite{yang2019xlnet}, which combines advantages of autoregressive and autoencoder models. Based on our task formulation, \textsc{Chime} is designed to maximize the probability $p(\bm{\hat{y}} |\bm{x}^{q}, \bm{x}^{r_{1}}\cdots \bm{x}^{r_{K}})$ of generating an answer given a question and its associated $K$ reviews in multi-passage review QA. The overall architecture of \textsc{Chime} is shown in Figure \ref{fig:Arch}. Given a question paired with $K$ text passages, we create $K$ training instances with each one consisting of the question, a text passage, and the best answer chosen by the helpfulness votes assigned by users. Each training instance is fed into an XLNet encoder to derive hidden representations, which will be used to update two memories. In particular, the \emph{context memory} is updated when seeing more text passages and the \emph{answer memory} is continuously refined with the answer generated from each (\emph{question, text passage}) pair. \textsc{Chime} has the following characteristics: (1) the use of a pre-trained XLNet as an encoder instead of traditional recurrent neural networks as the pre-trained LMs captures rich background knowledge and is more suitable for encoding semantic meanings of questions and review documents; (2) the proposal of the cross-passage \emph{context memory} mechanism to perform the reading of review passages in a sequential manner to deal with multiple text passages more effectively, which avoids the massive memory costs required to read all supporting passages in one go; (3) the use of the \emph{answer memory} to gradually refine the generated answer for a question after reading more text passages. Figure \ref{fig:Arch} shows the general architecture of \textsc{Chime}, which consists of three key components: the XLNet encoder for encoding a question, a review, and an answer, the cross-passage hierarchical memory mechanism, and the decoder for answer generation.

\paragraph{XLNet Encoder}
The XLNet Encoder in \textsc{Chime} is a vanilla XLNet encoder with special Seq2Seq masks introduced in UniLM \cite{dong2019unified}, which is essentially a concatenation of a standard pre-trained LM encoder and a pre-trained LM decoder. With the Seq2Seq masks, we are able to train an encoder for an encoder-decoder task. In specific, for each question paired with $K$ text passages, we create $K$ training instances with each one consisting of the triple \texttt{(question, passage, answer)}. We add the special token \texttt{[CLS]} at the beginning and insert \texttt{[SEP]} as a separator between every two elements in the triple and add another \texttt{[SEP]} at the end. In addition, we treat \texttt{([CLS] Question [SEP] Passage [SEP])} as Part 1 and \texttt{(Answer [SEP])} as Part 2. The Seq2Seq masks are designed in a way such that all tokens in Part 1 attend to each other, and tokens in Part 2 attend to any tokens in Part 1, but only preceding tokens in Part 2. Let $\bm{y}^{a_{g}}$ be the gold-standard answer selected for current training instance, $\bm{x}_{*}^{r_{k}}$ be the whole input sequence of instance $k$; $N_{x}$ be the length of $\bm{x}_{*}^{r_{k}}$, which keeps the same across all text passages; $d$ be the dimension of hidden size, and $H^{r_{k}}\in\mathbb{R}^{N_{x}\times d}$ be the contextual hidden states of the encoder:
\begin{linenomath}
  \begin{equation}
  \begin{split}
  \label{equa:xlnet encoder}
    & \bm{x}_{*}^{r_{k}} = \big[\mbox{[CLS]}\quad\bm{x}^{q}\quad\mbox{[SEP]}\quad\bm{x}^{r_{k}}\quad\mbox{[SEP]}\quad\bm{y^{a_{g}}}\quad\mbox{[SEP]}\big]\\
    & H^{r_{k}} = \mbox{XLNetEncoder}\big(E_{t}(x_{*}^{r_{k}})+E_{s}(x_{*}^{r_{k}})+E_{p}(x_{*}^{r_{k}})\big)\\
  \end{split}
  \end{equation}
\end{linenomath}
where $E_{t}(\cdot)$, $E_{s}(\cdot)$ and $E_{p}(\cdot)$ denote token embeddings, segment embeddings and position embeddings respectively. Here we use an interval segment embedding [$E_{t}^{A},E_{t}^{B},E_{t}^{A}$] to distinguish question, passage and answer other than the usual two-segment embedding in regular XLNet. As answers are only available during the training phase, training XLNet for the encoder-decoder task can be considered as fine-tuning pre-trained XLNet on our corpus in order to learn a better XLNet encoder.

\paragraph{Cross-passage hierarchical memory mechanism}
Hidden states of Part 1 and Part 2 are used to initialize and update \emph{context memory} and \emph{answer memory} respectively. Here the last \texttt{[SEP]} token in Part 1 is removed and added as the start token of Part 2 from this stage onwards for language generation purpose. Memory update is accomplished by taking a weighted aggregation of the previously retained memory and the current hidden state using a forget gate. The gate is obtained by using an MLP layer with a memory-specific Transformer encoder \cite{vaswani2017attention}, which is composed of a multi-head scaled dot product attention sublayer and a position-wise fully connected feed forward network sublayer. When receiving the hidden states derived from XLNet encoder, \textsc{Chime} first use the states of Part 1 to update \emph{context memory}, then hierarchically use the newly updated \emph{context memory} with the states of Part 2 to update \emph{answer memory}. Let $N_{S_{1}}$ and $N_{S_{2}}$ be the length of Part 1 and Part 2, respectively, which are kept the same across different text passages; $H_{c}^{r_{k}} \in\mathbb{R}^{N_{S_{1}}\times d}$ be the hidden states of the context part, which refers to the question and a text passage; $H_{a}^{r_{k}}\in\mathbb{R}^{N_{S_{2}}\times d}$ be the hidden states of the answer part; $M_{c}^{r_{k}}\in\mathbb{R}^{N_{S_{1}}\times d}$ and $M_{a}^{r_{k}}\in\mathbb{R}^{N_{S_{2}}\times d}$ be the updated \emph{context memory} and \emph{answer memory} respectively after reading $k$-th passage:
\begin{align*}
  \label{equa:memory update}
    Z^{r_{k}}_{c} &= \mbox{TransformerEncoder}(M_{c}^{r_{k-1}}, H_{c}^{r_{k}}) &  Z^{r_{k}}_{a} &= \mbox{TransformerEncoder}(H_{a}^{r_{k}}, M_{c}^{r_{k}}) \\
    G_{c}^{r_{k}} &= \sigma(W_{mc}^{r_{k}}M_{c}^{r_{k-1}} + W_{zc}^{r_{k}}Z^{r_{k}}_{c} + b_{c}^{r_{k}}) &
    G_{a}^{r_{k}} &= \sigma(W_{ha}^{r_{k}}H_{a}^{r_{k}} + W_{za}^{r_{k}}Z^{r_{k}}_{a} + b_{a}^{r_{k}}) \\
    M_{c}^{r_{k}} &= G_{c}^{r_{k}}M_{c}^{r_{k-1}}+(1-G_{c}^{r_{k}})H_{c}^{r_{k}} &
    M_{a}^{r_{k}} &= G_{a}^{r_{k}}H_{a}^{r_{k}}+(1-G_{a}^{r_{k}})M_{a}^{r_{k-1}} 
\end{align*}
where $Z^{r_{k}}_{c}\in\mathbb{R}^{N_{S_{1}}\times d}$ and $Z^{r_{k}}_{a}\in\mathbb{R}^{N_{S_{2}}\times d}$ denote the normalized attention output from the Transformer encoder, $G_{c}^{r_{k}}\in\mathbb{R}^{N_{S_{1}}\times d}_{[0,1]}$ and $G_{a}^{r_{k}}\in\mathbb{R}^{N_{S_{2}}\times d}_{[0,1]}$ denote the forget gate. $W_{mc}^{r_{k}}\in\mathbb{R}^{N_{S_{1}}\times d}$, $W_{zc}^{r_{k}}\in\mathbb{R}^{N_{S_{1}}\times d}$, $b_{c}^{r_{k}}\in\mathbb{R}^{N_{S_{1}}}$, $W_{ha}^{r_{k}}\in\mathbb{R}^{N_{S_{2}}\times d}$, $W_{za}^{r_{k}}\in\mathbb{R}^{N_{S_{2}}\times d}$ and $b_{a}^{r_{k}}\in\mathbb{R}^{N_{S_{2}}}$ are all trainable parameters. The two memories are initialized by taking the hidden states after reading the first review text passage of a question: $M_{c}^{r_{1}} = H_{c}^{r_{1}}, M_{a}^{r_{1}} = H_{a}^{r_{1}}$.

\paragraph{Decoder and Loss Function}
The answer probability $p(\bm{\hat{y}})$ over all $V$ tokens of the whole vocabulary is generated by adding a softmax layer on the top of the answer memory:
\begin{linenomath}
  \begin{equation}
  \label{equa:decoder}
    p(\bm{\hat{y}}) = \mbox{Softmax}(W_{ma}M_{a}^{r_{K}} + b_{a})\\
  \end{equation}
\end{linenomath}
where $W_{ma}\in\mathbb{R}^{d\times V}$ and $b_{a}\in\mathbb{R}^{V}$ are trainable. The training loss of each sample is the cross entropy loss of the predicted answer $\bm{\hat{y}}$
and gold-standard answer $\bm{y}$:
\begin{linenomath}
  \begin{equation}
  \label{equa:loss}
    L = -\frac{1}{N_{a}}\sum_{n=1}^{N_{a}}y_n\log \hat{y_n} \\
  \end{equation}
\end{linenomath}

\section{Experiments}
In this section, we first introduce the dataset used in our experiments, the baselines for comparison, and the evaluation metrics employed, followed by a discussion over the obtained results and a few examples generated using the different approaches presented.

\subsection{Settings}
\paragraph{Dataset} 
We built our dataset\footnote{Our dataset and codes are available at: \url{https://github.com/LuJunru/CHIME}.} from AmazonQA \cite{gupta2019amazonqa}. We only focused on more difficult `\emph{descriptive}' questions and filter out non-answerable or `yes/no' questions. We kept questions with 10 review snippets, sorting in descending order of relevance to the question. In the original dataset, 96\% of the answerable `\emph{descriptive}' questions are paired with 10 reviews. For each question, we only selected the best answer with the highest positive response rate. We further removed URL links from question, review, and answer text. The filtered dataset contains 365k samples in the training set, 47k samples in the validation set and \textcolor{black}{48k samples in the testing set}. We set the maximum tokenized lengths of questions, reviews, and answers to 40, 124, and 82, respectively, which cover 95\% of our samples.

\paragraph{Parameters setup}
The hidden size of BERT-base and XLNet-base is 768. The corresponding vocabulary sizes are 28,996 and 32,000. For \textsc{Chime}, the inner Transformer encoders are 1-block vanilla Transformer, which contains an 8-heads multi-head attention and a feed-forward network with 2048 inner state size. The optimizer of all neural baselines is AdamW \cite{loshchilov2018fixing} with $\beta1=0.9, \beta2=0.999$, and $\epsilon=1e-06$. Except for parameters of bias and layer normalization, all other training parameters are decayed with a rate of 0.95. The gradients of all parameters are clipped to the maximum norm 1.0. The learning rate is increased linearly from 0 to 1e-5 in the first 20\% total training steps and then linearly decreased to 0.

\paragraph{Baselines} 
We developed two heuristic baselines as well as three neural baselines:
\begin{itemize}
    \item \textbf{Random Sentence}. Given a question, select a random sentence from paired reviews as an answer.
    \item \textbf{Sentence Retrieval}. First, convert each question and each sentence of its paired reviews into sentence embeddings using BERT, then retrieve the sentences with the highest cosine similarity with the question as the selective answer. The sentence length of both heuristic baselines is 120.
    \item \textbf{BERT+summary}. Directly using BERT \cite{devlin2018bert} for generative QA is difficult since it is memory demanding to deal with multiple reviews in one go. We instead first generate an extractive summary of reviews using Textrank \cite{mihalcea2004textrank}, then feed a question and its associated review summary into BERT for answer generation.
    \item \textbf{XLNet+summary}. Although XLnet is theoretically capable of dealing with the text of unlimited length as it adopts the segmentation mechanism from Transformer XL \cite{dai2019transformer}, and could potentially process at once the concatenation of all the passages paired with a question, the computational requirements easily became rather prohibitive, and in practice is often not feasible to simultaneously deal with multiple long reviews with limited computational resources. Therefore, we take a similar summary-then-QA approach for XLNet.
    \item \textbf{XLNet+V-Net}. We follow the mutual verification mechanism proposed in V-Net \cite{wang2018multi} for answer post-processing. In particular, after XLNet generates candidate answers given individual reviews, mutual verification is conducted by calculating the average attention value of the current candidate answer with all the other answers. The one with the highest value is the final answer.
\end{itemize}

Due to the limitations of our computing resources, we have to use regular versions of large-scale pre-trained LMs and a subset of the original data. We use the BERT-base and the XLNet-base from Huggingface\footnote{\url{https://github.com/huggingface/transformers/blob/master/src/transformers}}. Both the neural baselines and our proposed \textsc{Chime} are trained with 25\% randomly selected data from our constructed dataset, which consists of 92k samples, comparable to popular large-scale datasets such as MS Marco (100k) \cite{Nguyen16} and HotpotQA (113k) \cite{yang2018hotpotqa}. For all neural models, we train for 3 epochs and use the beam search with size 3 over the best models to generate answers from decoder probability distributions. \textcolor{black}{In testing phase, 1k samples are extracted randomly for answer generation and evaluation.}

\paragraph{Metrics}
We use ROUGE-L \cite{lin04} and BLEU \cite{papineni02} to evaluate the lexical similarity between the gold-standard and the model generated answers. To measure the semantic similarity, we use BertScore\footnote{\url{https://github.com/Tiiiger/bert_score}} \cite{zhang2019bertscore}, which first computes the pairwise cosine similarity among all the tokens in the candidate and reference answers, and then greedily match them to get the highest similarity score for the sentence pair. BLEURT\footnote{\url{https://github.com/google-research/bleurt}} \cite{sellam2020bleurt} is a text generation quality evaluation framework that uses BLEU, ROUGE and BertScore and other indicators as multi-task joint training through fine-tuning BERT. We use BLEURT as a comprehensive metric to evaluate both the lexical and semantic similarities. A higher BLEURT score means that the generated sentence is both lexically and semantically closer to the ground truth. As each question is paired with multiple ground-truth answers, for BertScore and BleuRT, we finally consider the pair obtaining the maximum score.

\subsection{Results}
\begin{table}[ht!]
  \centering
  \begin{tabular}{cccccc}
  \toprule  
  Model & Bleu-1 & Bleu-2 & Rouge-L F1 & BertScore & BleuRT\\
  \midrule  
  \textbf{Heuristic Baselines}\\
  \hspace{0.5cm} Random Sentence & 25.189 & 8.996 & 15.669 & 0.103 & -1.043\\
  \hspace{0.6cm} Retrieval Sentence & 24.895 & 8.848 & 15.393 & 0.099 & -1.040\\
  \midrule  
  \hspace{-0.5cm} \textbf{Neural Baselines}\\
  \hspace{0.55cm} BERT + Summary & 31.404 & 14.494 & 16.856 & -0.027 & -1.376\\
  \hspace{0.6cm} XLNet + Summary & 32.037 & 14.018 & 20.484 & 0.162 & \textbf{-0.866}\\
  \hspace{0.05cm} XLNet + V-Net & 31.950 & 14.465 & 20.807 & 0.176 & -0.952\\
  \midrule  
  \hspace{-1.25cm} \textsc{Chime} & \textbf{33.103} & \textbf{14.947} & \textbf{21.512} & \textbf{0.185} & -0.949\\
  \hspace{-0.95cm} \textsc{Chime}{-c} & 29.552 & 14.202 & 20.831 & 0.174 & -0.982\\
  \hspace{-0.95cm} \textsc{Chime}{-a} & 31.361 & 14.142 & 20.988 & 0.177 & -0.976\\
  \bottomrule 
  \end{tabular}
  \caption{Evaluation results of \textsc{Chime}s and baselines. The answers generated by \textsc{Chime}s are superior in terms of lexical and semantics evaluations. \textsc{Chime}{-c} removes the \emph{context memory} and makes use of just the \emph{answer memory}, in which the \emph{answer memory} is updated not by \emph{context memory} but by current context hidden states. In contrast, \textsc{Chime}{-a} removes the \emph{answer memory} and makes use of just the \emph{context memory}, in which we remove the MLP sublayer for \emph{answer memory} and directly feed the output of middle transformer encoder to the final decoder as shown in Figure \ref{fig:Arch}.}
  \label{tab:full}
\end{table}

Table \ref{tab:full} reports the evaluations of 1k selective samples from the testing set. The answers generated by \textsc{Chime} exhibit an overall improved quality reflected by lexical and semantic evaluations outperforming all baselines. This validates the efficacy of combining the \textit{context} and the \textit{answer memory} to generate coherent answers when processing multiple passages, containing possibly contradictory opinions. \textsc{Chime}{-c} is an ablated version of \textsc{Chime} that only uses the \emph{answer memory}, which is updated without the link from the \emph{context memory} $M_{c}^{r_{k}}$ but using the current context hidden states $H_{c}^{r_{k}}$. The comparison of \textsc{Chime}{-c} with \textsc{Chime} demonstrates the importance of the cross-passage evidence collection. Similarly, \textsc{Chime}{-a} is another ablated version that makes use of the only \emph{context memory}, in which we link $Z^{r_{k}}_{a}$ from the \emph{answer memory}'s encoder for the final decoding. The performance gap between \textsc{Chime}{-a} and \textsc{Chime} corroborate the relevance of a gradual answer refinement.

\subsection{Qualitative analysis}
\begin{table}[h!]
  \centering
  \begin{tabular}{l}
  \toprule  
  \textbf{Question}: \tabincell{l}{Can I set it for 30 seconds on the memory function, and then can it will automatically \\reset to the 30 seconds continually?}\\
  \textbf{Answer 1}: Yes, you can reset the times you set and reuse them.\\
  \textbf{Answer 2}: No...thats why it sucks...\\
  \textbf{Answer 3}: Do not know\\
  \textbf{Answer 4}: yes. the timer has 4 different timer settings. timer 1 is the regular one that gets used the \\most. but, timers 2, 3 and 4 can be programmed in memory so that you can recall the time every \\time you want to use it. see my other post, though. the first time this gets knocked off the counter, \\it’s history.\\
  \midrule  
  \textbf{Random Sentence}: the only way to reset to the memorized time is to hit the ”memory 2-3-4” button \\4x to cycle through all 4 timers and get back to the one you are using.\\
  \textbf{Retrieval Sentence}: but, and here’s the major flaw, once the timer counts down to 0:00, beeps, and \\you hit the stop button, the time remains at 0:00.\\
  \textbf{BERT + Summary}: i have not have a little good. i don’t have a little light. i’m not sure, but you can.\\
  \textbf{XLNet + Summary}: i have no problem with this. it works for 30 mins, but the timer will reset to 0, \\but not to a point where you can adjust your timer.\\
  \textbf{XLNet + V-Net}: yes, it will reset to 30 seconds, but i don't know if it would work for 30 minutes \\or not? i have no idea about this issue but.\\
  \textbf{\textsc{Chime}}: yes you can set it for 30 seconds. i have set mine for 30 minutes and have had no problems \\with that timer going off and running again.\\
  \bottomrule 
  \end{tabular}
  \caption{Comparison of example answers. \textsc{Chime} outperforms all the five baselines in generating more reliable answer. XLNets provide readable but noisy, incomplete and ill-understandable answers. All other baselines including the BERT one are not able to generate readable answers.}
  \label{tab:compare1}
\end{table}

As a case study, we analyze the example reported in Figure \ref{fig:illus}. We first compare the quality of the answers generated by different models and then illustrate a breakdown of the CHIME’s generative process when iteratively reading different reviews. The gradual generative process provides some explicit interpretability of cross-passage evidence collection and sequential answer refinement.

In Table \ref{tab:compare1} we compare a few answers generated using different models\footnote{More example outputs are presented in Appendix.}. Answers returned by either randomly selecting a sentence from review text passages or by retrieving a sentence from passages which is most similar to a given question are clearly not directly addressing the question. The poor quality of the answer returned by the BERT$+$Summary model, off-topic and ill-grammatical structure, shows the limitation of simply using the out-of-the-box BERT in text generation. The XLNets are able to generate some reliable answers, which is much better than the BERT$+$Summary. However, the two XLNet models mistakenly uses "30 mins" to replace the key term "30 seconds", which weakens the credibility of the answer. Compared with XLNets, the \textsc{Chime} generates syntactically well-formed answer with better coherency and fluency.

\begin{figure}[ht!]
\centering
\includegraphics[scale=0.24]{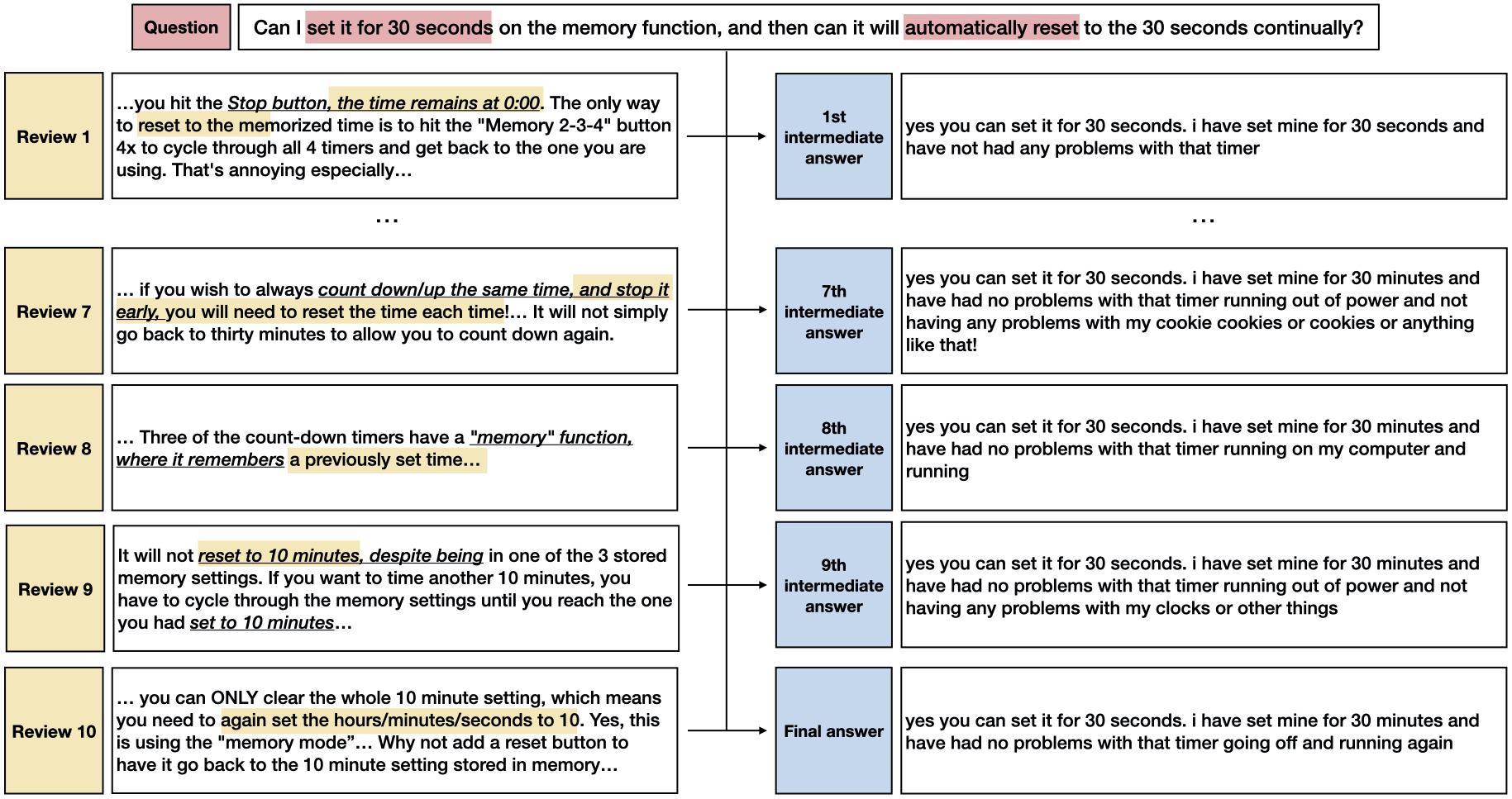}
\caption{A breakdown of \textsc{Chime}'s generative process. The example question (the top box), the paired reviews (left panel), and the intermediate answers (right panel) after gradually reading the corresponding reviews. The major points of the question and reviews are highlighted with colors, and the \emph{italic text} marked with \underline{underline} is the content most concerned by the forget gate. Given new reviews, the very first generated simple answer becomes complicated and full of noise, but finally converges to the most prominent opinions and facts relevant to the question.}
\label{fig:break}
\end{figure}

Figure \ref{fig:break} shows a breakdown of \textsc{Chime}'s generative process. The question-related content highlighted with colors is highly likely the major concerning part that the forget gate believes to memorise. The intermediate answers reported show that \textsc{Chime} has locked the answer to the first sub-question from the beginning. But for the second sub-question, as the 8th answer shows, \textsc{Chime} was also misled by other unimportant information. The final answer is eventually a synthesis of the prominent opinions encountered, summarised in a few concise phrases.

\section{Conclusions}

In this paper, we have proposed \textsc{Chime}, a cross-passage hierarchical memory network for multi-passage generative review QA. It is built on the XLNet generator \cite{yang2019xlnet} by adding a memory module consisting of a \textit{context} and a \textit{answer memory} which guarantees a more accurate refining process for cross-passage evidence collection and answer generation. The sequential process adopted in \textsc{Chime} makes it possible to elaborate longer text passages and some straightforward interpretability. We have assessed experimentally a significant quality improvement using different state-of-the-art metrics to measure the lexical and semantic coherence of the generated text. We plan to further extend \textsc{Chime} to model with multiple ground truth simultaneously and leverage the available product attributes.

\section{Acknowledgements}
We would like to thank Mansi Gupta and her co-authors for sharing the test set of AmazonQA \cite{gupta2019amazonqa} with us. This work was funded in part by EPSRC (grant no. EP/T017112/1), EU-H2020 (grant no. 794196), the Natural Science Foundation of China (61976066), and the Fundamental Research Fund for the Central Universities
(2019GA35).

\bibliographystyle{coling}
\bibliography{chime}

\begin{thebibliography}{}

\bibitem[\protect\citename{Chen \bgroup et al.\egroup
  }2019]{chen2019bidirectional}
Yu~Chen, Lingfei Wu, and Mohammed~J Zaki.
\newblock 2019.
\newblock Bidirectional attentive memory networks for question answering over
  knowledge bases.
\newblock {\em arXiv preprint arXiv:1903.02188}.

\bibitem[\protect\citename{Dai \bgroup et al.\egroup }2019]{dai2019transformer}
Zihang Dai, Zhilin Yang, Yiming Yang, Jaime Carbonell, Quoc~V Le, and Ruslan
  Salakhutdinov.
\newblock 2019.
\newblock Transformer-xl: Attentive language models beyond a fixed-length
  context.
\newblock {\em arXiv preprint arXiv:1901.02860}.

\bibitem[\protect\citename{Devlin \bgroup et al.\egroup }2018]{devlin2018bert}
Jacob Devlin, Ming-Wei Chang, Kenton Lee, and Kristina Toutanova.
\newblock 2018.
\newblock Bert: Pre-training of deep bidirectional transformers for language
  understanding.
\newblock {\em arXiv preprint arXiv:1810.04805}.

\bibitem[\protect\citename{Dong \bgroup et al.\egroup }2019]{dong2019unified}
Li~Dong, Nan Yang, Wenhui Wang, Furu Wei, Xiaodong Liu, Yu~Wang, Jianfeng Gao,
  Ming Zhou, and Hsiao-Wuen Hon.
\newblock 2019.
\newblock Unified language model pre-training for natural language
  understanding and generation.
\newblock In {\em Advances in Neural Information Processing Systems}, pages
  13042--13054.

\bibitem[\protect\citename{Fan \bgroup et al.\egroup }2018]{fan2018convolution}
Chuang Fan, Qinghong Gao, Jiachen Du, Lin Gui, Ruifeng Xu, and Kam-Fai Wong.
\newblock 2018.
\newblock Convolution-based memory network for aspect-based sentiment analysis.
\newblock In {\em The 41st International ACM SIGIR Conference on Research \&
  Development in Information Retrieval}, pages 1161--1164.

\bibitem[\protect\citename{Fu \bgroup et al.\egroup }2020]{fu2020recurrent}
Jinlan Fu, Yi~Li, Qi~Zhang, Qinzhuo Wu, Renfeng Ma, Xuanjing Huang, and Yu-Gang
  Jiang.
\newblock 2020.
\newblock Recurrent memory reasoning network for expert finding in community
  question answering.
\newblock In {\em Proceedings of the 13th International Conference on Web
  Search and Data Mining}, pages 187--195.

\bibitem[\protect\citename{Gao \bgroup et al.\egroup }2019]{gao2019product}
Shen Gao, Zhaochun Ren, Yihong Zhao, Dongyan Zhao, Dawei Yin, and Rui Yan.
\newblock 2019.
\newblock Product-aware answer generation in e-commerce question-answering.
\newblock In {\em Proceedings of the Twelfth ACM International Conference on
  Web Search and Data Mining}, pages 429--437.

\bibitem[\protect\citename{Gui \bgroup et al.\egroup }2017]{gui2017question}
Lin Gui, Jiannan Hu, Yulan He, Ruifeng Xu, Qin Lu, and Jiachen Du.
\newblock 2017.
\newblock A question answering approach to emotion cause extraction.
\newblock {\em arXiv preprint arXiv:1708.05482}.

\bibitem[\protect\citename{Gupta \bgroup et al.\egroup
  }2019]{gupta2019amazonqa}
Mansi Gupta, Nitish Kulkarni, Raghuveer Chanda, Anirudha Rayasam, and Zachary~C
  Lipton.
\newblock 2019.
\newblock Amazonqa: a review-based question answering task.
\newblock {\em arXiv preprint arXiv:1908.04364}.

\bibitem[\protect\citename{Guthrie and Mosenthal}1987]{1987Literacy}
John~T. Guthrie and Peter Mosenthal.
\newblock 1987.
\newblock Literacy as multidimensional: Locating information and reading
  comprehension.
\newblock {\em Educational Psychologist}, 22(3-4):279--297.

\bibitem[\protect\citename{Kim \bgroup et al.\egroup }2018]{kim2018abstractive}
Byeongchang Kim, Hyunwoo Kim, and Gunhee Kim.
\newblock 2018.
\newblock Abstractive summarization of reddit posts with multi-level memory
  networks.
\newblock {\em arXiv preprint arXiv:1811.00783}.

\bibitem[\protect\citename{Kumar \bgroup et al.\egroup }2016]{kumar2016ask}
Ankit Kumar, Ozan Irsoy, Peter Ondruska, Mohit Iyyer, James Bradbury, Ishaan
  Gulrajani, Victor Zhong, Romain Paulus, and Richard Socher.
\newblock 2016.
\newblock Ask me anything: Dynamic memory networks for natural language
  processing.
\newblock In {\em International conference on machine learning}, pages
  1378--1387.

\bibitem[\protect\citename{Lin}2004]{lin04}
Chin-Yew Lin.
\newblock 2004.
\newblock {ROUGE}: A package for automatic evaluation of summaries.
\newblock In {\em Text Summarization Branches Out}, pages 74--81, Barcelona,
  Spain, July. Association for Computational Linguistics.

\bibitem[\protect\citename{Loshchilov and Hutter}2018]{loshchilov2018fixing}
Ilya Loshchilov and Frank Hutter.
\newblock 2018.
\newblock Fixing weight decay regularization in adam.

\bibitem[\protect\citename{Maruf and Haffari}2017]{maruf2017document}
Sameen Maruf and Gholamreza Haffari.
\newblock 2017.
\newblock Document context neural machine translation with memory networks.
\newblock {\em arXiv preprint arXiv:1711.03688}.

\bibitem[\protect\citename{McAuley and Yang}2016]{mcauley2016addressing}
Julian McAuley and Alex Yang.
\newblock 2016.
\newblock Addressing complex and subjective product-related queries with
  customer reviews.
\newblock In {\em Proceedings of the 25th International Conference on World
  Wide Web}, pages 625--635.

\bibitem[\protect\citename{Mihalcea and Tarau}2004]{mihalcea2004textrank}
Rada Mihalcea and Paul Tarau.
\newblock 2004.
\newblock Textrank: Bringing order into text.
\newblock In {\em Proceedings of the 2004 conference on empirical methods in
  natural language processing}, pages 404--411.

\bibitem[\protect\citename{Nguyen \bgroup et al.\egroup }2016]{Nguyen16}
Tri Nguyen, Mir Rosenberg, Xia Song, Jianfeng Gao, Saurabh Tiwary, Rangan
  Majumder, and Li~Deng.
\newblock 2016.
\newblock {MS} {MARCO:} {A} human generated machine reading comprehension
  dataset.
\newblock In Tarek~Richard Besold, Antoine Bordes, Artur~S. d'Avila Garcez, and
  Greg Wayne, editors, {\em Proceedings of the Workshop on Cognitive
  Computation: Integrating neural and symbolic approaches 2016 co-located with
  the 30th Annual Conference on Neural Information Processing Systems {(NIPS}
  2016), Barcelona, Spain, December 9, 2016}, volume 1773 of {\em {CEUR}
  Workshop Proceedings}.

\bibitem[\protect\citename{Nishida \bgroup et al.\egroup }2019]{nishida19}
Kyosuke Nishida, Itsumi Saito, Kosuke Nishida, Kazutoshi Shinoda, Atsushi
  Otsuka, Hisako Asano, and Junji Tomita.
\newblock 2019.
\newblock Multi-style generative reading comprehension.
\newblock In {\em Proceedings of the 57th Annual Meeting of the Association for
  Computational Linguistics}, pages 2273--2284, Florence, Italy, July.
  Association for Computational Linguistics.

\bibitem[\protect\citename{Papineni \bgroup et al.\egroup }2002]{papineni02}
Kishore Papineni, Salim Roukos, Todd Ward, and Wei-Jing Zhu.
\newblock 2002.
\newblock {B}leu: a method for automatic evaluation of machine translation.
\newblock In {\em Proceedings of the 40th Annual Meeting of the Association for
  Computational Linguistics}, pages 311--318, Philadelphia, Pennsylvania, USA,
  July. Association for Computational Linguistics.

\bibitem[\protect\citename{Petroni \bgroup et al.\egroup
  }2019]{petroni2019language}
Fabio Petroni, Tim Rockt{\"a}schel, Patrick Lewis, Anton Bakhtin, Yuxiang Wu,
  Alexander~H Miller, and Sebastian Riedel.
\newblock 2019.
\newblock Language models as knowledge bases?
\newblock {\em arXiv preprint arXiv:1909.01066}.

\bibitem[\protect\citename{Raffel \bgroup et al.\egroup
  }2019]{raffel2019exploring}
Colin Raffel, Noam Shazeer, Adam Roberts, Katherine Lee, Sharan Narang, Michael
  Matena, Yanqi Zhou, Wei Li, and Peter~J Liu.
\newblock 2019.
\newblock Exploring the limits of transfer learning with a unified text-to-text
  transformer.
\newblock {\em arXiv preprint arXiv:1910.10683}.

\bibitem[\protect\citename{Rajpurkar \bgroup et al.\egroup
  }2016]{rajpurkar2016squad}
Pranav Rajpurkar, Jian Zhang, Konstantin Lopyrev, and Percy Liang.
\newblock 2016.
\newblock Squad: 100,000+ questions for machine comprehension of text.
\newblock {\em arXiv preprint arXiv:1606.05250}.

\bibitem[\protect\citename{Robertson and
  Zaragoza}2009]{robertson2009probabilistic}
Stephen Robertson and Hugo Zaragoza.
\newblock 2009.
\newblock {\em The probabilistic relevance framework: BM25 and beyond}.
\newblock Now Publishers Inc.

\bibitem[\protect\citename{Sellam \bgroup et al.\egroup
  }2020]{sellam2020bleurt}
Thibault Sellam, Dipanjan Das, and Ankur~P Parikh.
\newblock 2020.
\newblock Bleurt: Learning robust metrics for text generation.
\newblock {\em arXiv preprint arXiv:2004.04696}.

\bibitem[\protect\citename{Sukhbaatar \bgroup et al.\egroup
  }2015]{sukhbaatar2015end}
Sainbayar Sukhbaatar, Jason Weston, Rob Fergus, et~al.
\newblock 2015.
\newblock End-to-end memory networks.
\newblock In {\em Advances in neural information processing systems}, pages
  2440--2448.

\bibitem[\protect\citename{Tan \bgroup et al.\egroup }2018]{Tan18}
Chuanqi Tan, Furu Wei, Nan Yang, Bowen Du, Weifeng Lv, and Ming Zhou.
\newblock 2018.
\newblock S-net: From answer extraction to answer synthesis for machine reading
  comprehension.
\newblock In {\em AAAI}.

\bibitem[\protect\citename{Trischler \bgroup et al.\egroup
  }2016]{trischler2016newsqa}
Adam Trischler, Tong Wang, Xingdi Yuan, Justin Harris, Alessandro Sordoni,
  Philip Bachman, and Kaheer Suleman.
\newblock 2016.
\newblock Newsqa: A machine comprehension dataset.
\newblock {\em arXiv preprint arXiv:1611.09830}.

\bibitem[\protect\citename{Vaswani \bgroup et al.\egroup
  }2017]{vaswani2017attention}
Ashish Vaswani, Noam Shazeer, Niki Parmar, Jakob Uszkoreit, Llion Jones,
  Aidan~N Gomez, {\L}ukasz Kaiser, and Illia Polosukhin.
\newblock 2017.
\newblock Attention is all you need.
\newblock In {\em Advances in neural information processing systems}, pages
  5998--6008.

\bibitem[\protect\citename{Wan and McAuley}2016]{wan2016modeling}
Mengting Wan and Julian McAuley.
\newblock 2016.
\newblock Modeling ambiguity, subjectivity, and diverging viewpoints in opinion
  question answering systems.
\newblock In {\em 2016 IEEE 16th international conference on data mining
  (ICDM)}, pages 489--498. IEEE.

\bibitem[\protect\citename{Wang \bgroup et al.\egroup }2017]{wang17}
Wenhui Wang, Nan Yang, Furu Wei, Baobao Chang, and Ming Zhou.
\newblock 2017.
\newblock Gated self-matching networks for reading comprehension and question
  answering.
\newblock In {\em Proceedings of the 55th Annual Meeting of the Association for
  Computational Linguistics (Volume 1: Long Papers)}, Vancouver, Canada, July.

\bibitem[\protect\citename{Wang \bgroup et al.\egroup }2018]{wang2018multi}
Yizhong Wang, Kai Liu, Jing Liu, Wei He, Yajuan Lyu, Hua Wu, Sujian Li, and
  Haifeng Wang.
\newblock 2018.
\newblock Multi-passage machine reading comprehension with cross-passage answer
  verification.
\newblock {\em arXiv preprint arXiv:1805.02220}.

\bibitem[\protect\citename{Wang \bgroup et al.\egroup }2019]{wang2019multi}
Zhiguo Wang, Patrick Ng, Xiaofei Ma, Ramesh Nallapati, and Bing Xiang.
\newblock 2019.
\newblock Multi-passage bert: A globally normalized bert model for open-domain
  question answering.
\newblock {\em arXiv preprint arXiv:1908.08167}.

\bibitem[\protect\citename{Weston \bgroup et al.\egroup
  }2014]{weston2014memory}
Jason Weston, Sumit Chopra, and Antoine Bordes.
\newblock 2014.
\newblock Memory networks.
\newblock {\em arXiv preprint arXiv:1410.3916}.

\bibitem[\protect\citename{Xiong \bgroup et al.\egroup }2016]{xiong2016dynamic}
Caiming Xiong, Stephen Merity, and Richard Socher.
\newblock 2016.
\newblock Dynamic memory networks for visual and textual question answering.
\newblock In {\em International conference on machine learning}, pages
  2397--2406.

\bibitem[\protect\citename{Xu \bgroup et al.\egroup }2019a]{xu2019bert}
Hu~Xu, Bing Liu, Lei Shu, and Philip~S Yu.
\newblock 2019a.
\newblock Bert post-training for review reading comprehension and aspect-based
  sentiment analysis.
\newblock {\em arXiv preprint arXiv:1904.02232}.

\bibitem[\protect\citename{Xu \bgroup et al.\egroup }2019b]{xu2019enhancing}
Kun Xu, Yuxuan Lai, Yansong Feng, and Zhiguo Wang.
\newblock 2019b.
\newblock Enhancing key-value memory neural networks for knowledge based
  question answering.
\newblock In {\em Proceedings of the 2019 Conference of the North American
  Chapter of the Association for Computational Linguistics: Human Language
  Technologies, Volume 1 (Long and Short Papers)}, pages 2937--2947.

\bibitem[\protect\citename{Yang \bgroup et al.\egroup }2018]{yang2018hotpotqa}
Zhilin Yang, Peng Qi, Saizheng Zhang, Yoshua Bengio, William~W Cohen, Ruslan
  Salakhutdinov, and Christopher~D Manning.
\newblock 2018.
\newblock Hotpotqa: A dataset for diverse, explainable multi-hop question
  answering.
\newblock {\em arXiv preprint arXiv:1809.09600}.

\bibitem[\protect\citename{Yang \bgroup et al.\egroup }2019]{yang2019xlnet}
Zhilin Yang, Zihang Dai, Yiming Yang, Jaime Carbonell, Russ~R Salakhutdinov,
  and Quoc~V Le.
\newblock 2019.
\newblock Xlnet: Generalized autoregressive pretraining for language
  understanding.
\newblock In {\em Advances in neural information processing systems}, pages
  5754--5764.

\bibitem[\protect\citename{Yu and Lam}2018]{yu2018aware}
Qian Yu and Wai Lam.
\newblock 2018.
\newblock Review-aware answer prediction for product-related questions
  incorporating aspects.
\newblock In {\em Proceedings of the Eleventh ACM International Conference on
  Web Search and Data Mining}, pages 691--699.

\bibitem[\protect\citename{Zhang \bgroup et al.\egroup
  }2019]{zhang2019bertscore}
Tianyi Zhang, Varsha Kishore, Felix Wu, Kilian~Q Weinberger, and Yoav Artzi.
\newblock 2019.
\newblock Bertscore: Evaluating text generation with bert.
\newblock {\em arXiv preprint arXiv:1904.09675}.

\bibitem[\protect\citename{Zheng \bgroup et al.\egroup }2019]{2019Human}
Yukun Zheng, Jiaxin Mao, Yiqun Liu, Zixin Ye, and Shaoping Ma.
\newblock 2019.
\newblock Human behavior inspired machine reading comprehension.
\newblock In {\em the 42nd International ACM SIGIR Conference}.

\end{thebibliography}

\appendixtitleon
\appendixtitletocon

\newpage
\begin{appendices}
\renewcommand\thetable{\Alph{section}\arabic{table}}

\section{More Example Outputs Generated by Various Models}
\label{sec:suppex}
\setcounter{table}{0}

\begin{table}[h!]
  \centering
  \begin{tabular}{l}
  \toprule  
  \textbf{(a) Question}: does this come with the straps?/\\
  \textbf{Answer 1}: no it doesn't sadly.\\
  \textbf{Answer 2}: no. you have to buy separate.\\
  \textbf{Answer 3}: it comes with the white cotton rope ending in knots typical of mayan hammocks \\but no tree hugger straps or hammock frame.\\
  \textbf{Answer 4}: no it does not come with the straps or rope. i advise you to purchase parachord with it.\\
  \textbf{XLNet + Summary}: yes, it comes with the straps and the hammock is included with it! it comes \\in two pairs of harnesses that are attached to each other, so you can use them on your porch or \\your bed, or just use them on a bed or bed\\
  \textbf{XLNet + V-Net}: yes it comes with the straps and the sizing is very good and very easy to do!\\
  \textbf{\textsc{Chime}}: yes it does. i love this hammock and love the quality of the straps\\
  \midrule  
  \textbf{(b) Question}: has anyone tried baking with this?\\
  \textbf{Answer 1}: yes you can! designer whey also has recipes that you can make, like cookies, waffles \\or pancakes and muffins!\\
  \textbf{XLNet + Summary}: i have never tried it with my own breads but i have had a few good ones and \\they were great!\\
  \textbf{XLNet + V-Net}: i haven't tried baking with it, but i've been using this for a couple of weeks \\now and\\
  \textbf{\textsc{Chime}}: i haven't tried baking with this, but the flavor is very good and the flavor is very nice \\and good!\\
  \midrule  
  \textbf{(c) Question}: can you put coals on the lid, like you can with some lodge models?\\
  \textbf{Answer 1}: yes you can,\\
  \textbf{Answer 2}: yes. i've done it many times. make sure you get one of those lid-lifters though so you \\don't pour ash or coals into the pot.\\
  \textbf{Answer 3}: yes, this dutch oven is specifically designed with a tall lip to cup the coals on top of \\the lid to allow for baking. this has worked great for me in making cobbler and biscuits.\\
  \textbf{Answer 4}: never heard of coals on the lid, sorry. it makes a great chili over an open pit bbq. we \\hang it over a pit in the winter out back and make stews, chili's and pot luck. makes for a fun \\time especially in a snow storm!\\
  \textbf{Answer 5}: shouldn't be a problem. lid fits with good seal so no coals or ash should spill inside.\\
  \textbf{XLNet + Summary}: yes, you can. i don't know if it is possible to put coals on the top of the lid \\and then place them on top of that\\
  \textbf{XLNet + V-Net}: yes, you can put coals on the lid, but i don't think you can put coals inside the top, \\so i don've had any problems with this oven and have not tried any of these models with any \\problem with mine\\
  \textbf{\textsc{Chime}}: yes, you can put coals on the lid, but i don't think you can put any coal in this one, as the \\lid is not designed for that type of use\\
  \midrule  
  \textbf{(d) Question}: in what stores are these sold?\\
  \textbf{Answer 1}: target and target.com is where i've bought 3 pairs over time.\\
  \textbf{Answer 2}: different sizes different stores.\\
  \textbf{XLNet + Summary}: i don't know. i have never used them in my life, but they are very good for the \\foot, but it is not as good as the ones that are sold at stores\\
  \textbf{XLNet + V-Net}: i don't know. i've been using them for a couple of years and it works great! they are \\very good and very good!\\
  \textbf{\textsc{Chime}}: amazon has them in their store. i have had them for over 2 years and have never had any \\problems with it working for my feet, especially when\\
  \bottomrule 
  \end{tabular}
  \caption{Example outputs to compare the quality of \textsc{Chime} answers and XLNets answers.}
  \label{tab:more}
\end{table}

\newpage
\begin{table}[ht]
  \centering
  \begin{tabular}{l}
  \toprule  
  \textbf{(e) Question}: is this set compatible with the new canon powershot sx50 hs?\\
  \textbf{Answer 1}: yes james, as long as you decide what mm size you want to use! i prefered the 58mm \\adapter ring, this info is available in the canon online operaters manual! just make sure you specify \\the camera number (sx50 hs) when ordering any after market devices!! ps: you may already have \\52mm (or other) lenses from another camera,\\
  \textbf{Answer 2}: yes\\
  \textbf{Answer 3}: i don't know as i have the sx40. as far as the sx40 i can say that everything fits and works \\with the camera very nice. i've now used everything. the lens cover is much much better than the one \\that came with the camera. the little case that houses the lenses is very nice. once you screw the \\adapter ring onto the end\\
  \textbf{Answer 4}: i bought it for my canon sx40 and it works perfectly, but i'm not sure if the sx50 lens \\diameter is larger. this set does come with an adapter ring that fit snugly on my canon sx40 which \\allows placement of the filters and you can also stack the filters and/or add the lens hood. i mainly just \\use the lens hood\\
  \textbf{Answer 5}: i just recieved this in the mail and the adapter does not fit my sx50??\\
  \textbf{Answer 6}: hi everyone! the sx40 and sx50 have very similar dimensions and many people were able \\to make this ring fit on the sx50, however please know that there is an adapter ring designed \\specifically for use with the sx50. you can purchase this exact kit with the sx50 ring by following this \\link essential accessory kit for canon powershot sx\\
  \textbf{XLNet + Summary}: yes, it is compatible with the new canon powershot and the new canon power \\shot camera, but it is compatible with all other cameras that have a different camera and\\
  \textbf{XLNet + V-Net}: yes it is compatible with the sx40 sx20\\
  \textbf{CHIME}: yes, it is compatible with the new canon sx50\\
  \midrule  
  \textbf{(f) Question}: does it work on an iphone 4 or less?\\
  \textbf{Answer 1}: yes, it will work. it will not work with any higher versions only lower.\\
  \textbf{Answer 2}: yes as the plugs are the same.. but to tell you the truth i wouldn't bother buying one if i \\had my time over. not long after i ordered my dock sonos released an uprade, "play from this phone" \\which basically meant you didn't have to have your computer on the listen to music..i only bought \\the dock for this very purpose\\
  \textbf{XLNet + Summary}: yes, it does. i have a iphone 4 and it works fine with my 4 year old\\
  \textbf{XLNet + V-Net}: yes it works on my iphone 4 and it works great! i have a sonos and have had no \\problems with the device running with my sono 5, but the cord is not very long, so if you are looking \\for an older model you should be able to find one that will do that\\
  \textbf{\textsc{Chime}}: yes it works on my iphone 4 and yes, works with my sonos\\
  \midrule  
  \textbf{(g) Question}: how is this mounted on a deck railing that is 1 1/4" wide?\\
  \textbf{Answer 1}: the hanger just sits over the railing but it also has the side piece that a lot of other planters \\don't have. that extra piece rests against the railing at a ninety degree angle and keeps the planter from \\tipping forward. it's great...i love mine!\\
  \textbf{Answer 2}: it has adjustable brackets that attach on each end\\
  \textbf{Answer 3}: it is adjustable\\
  \textbf{Answer 4}: it has adjustable brackets. it works on my pull fence quite well.\\
  \textbf{XLNet + Summary}: it's a 2" wide. it's a 2 1/4" long, but i would say that this is the best way to mount \\the rack on any railing, so if you are looking for an adjustable bracket for your deck or balcony or window\\
  \textbf{XLNet + V-Net}: i've used mine on a deck with no problems. i've had no problems on my deck railing, \\but the rail is not sturdy enough to hold up to that height of my house\\
  \textbf{\textsc{Chime}}: i have a deck railing that is 1 1/4" wide. i would recommend using a bracket to hold the box in \\place, but if you are going to hang your deck railing over\\
  \bottomrule 
  \end{tabular}
  \caption{Example outputs to compare the quality of \textsc{Chime} answers and XLNets answers.}
  \label{tab:more}
\end{table}

\end{appendices}
\end{document}